\documentclass{article}

\usepackage{amsmath,amssymb,amsfonts}
\usepackage{graphicx}
\usepackage{textcomp}
\usepackage{xcolor}
\usepackage{stfloats}
\usepackage{subcaption}
\usepackage{url}
\usepackage{verbatim}
\usepackage{layouts}
\usepackage{adjustbox}
\usepackage{float}
\usepackage{xspace}
\usepackage{multirow}
\usepackage{pbox}
\usepackage{multicol}
\usepackage{makecell}
\usepackage{array}
\usepackage{colortbl}
\usepackage{booktabs}
\usepackage{tabularx}
\usepackage[absolute,overlay]{textpos}
\usepackage{siunitx}
\usepackage[accsupp]{axessibility}  %
\usepackage{adjustbox}
\usepackage{tabularx}
\usepackage{enumitem}
\usepackage[preprint]{corl_2026} %

\title{Worth Remembering: Surprise-Gated Robot Episodic Memory}

\author{Nicolas Gorlo$^1$ \qquad Derek K. Wise$^{2}$ \qquad Alberto Speranzon$^{2}$ \qquad Luca Carlone$^1$\thanks{Luca holds concurrent appointments as a faculty at the Massachusetts Institute of Technology and as an Amazon Scholar. This paper describes work performed at MIT and is not associated with Amazon.}\\
$^1$Massachusetts Institute of Technology, $^2$Lockheed Martin}

\makeatletter
\def\@gobbletrailingpunct{%
  \@ifnextchar.{\@gobble\@gobbletrailingpunct}{%
    \@ifnextchar,{\@gobble\@gobbletrailingpunct}{%
      \@ifnextchar\space{\@gobble\@gobbletrailingpunct}{}%
  }%
  }%
}
\newcommand{\linkToPdf}[1]{\@gobbletrailingpunct}
\newcommand{\linkToPpt}[1]{\@gobbletrailingpunct}
\newcommand{\linkToCode}[1]{\@gobbletrailingpunct}
\newcommand{\linkToWeb}[1]{\@gobbletrailingpunct}
\newcommand{\linkToVideo}[1]{\@gobbletrailingpunct}
\newcommand{\linkToMedia}[1]{\@gobbletrailingpunct}
\newcommand{\award}[1]{\@gobbletrailingpunct}
\makeatother

\usepackage[capitalise]{cleveref}

\crefname{appsec}{Appendix}{Appendices}
\Crefname{appsec}{Appendix}{Appendices}

\begin{document}
\maketitle

\begin{abstract}
    Robots solving generalist tasks need to be able to ground instructions in their past experience, since  humans may refer to notable past events when giving a task (\textit{e.g.}, ``Take me to where the chemical spill happened yesterday''). Since memory limits make storing all past events infeasible, long-term robot memory must be selective,  ideally retaining only those episodes with high utility for future tasks. However, future tasks are not typically given a priori for generalist robots. To select generically useful memories, we propose \emph{Bayesian surprise as a gating mechanism for memory formation.} We present an approach to compute surprise in a semantically rich deployment-agnostic latent space provided by V-JEPA-2. Using our gated episodic memory to augment 4D scene graph-based spatial memory, we show a consistent improvement over state-of-the-art benchmarks in robot question answering, outperforming prior robot memory methods by $\geq12\%$ for temporal, spatial, and binary questions, and surpassing the performance of supervised and non-causal methods with an unsupervised causal method in event segmentation tasks.
\end{abstract}


\begin{figure}[H]
    \captionsetup{type=figure}
    \includegraphics[width=\textwidth]{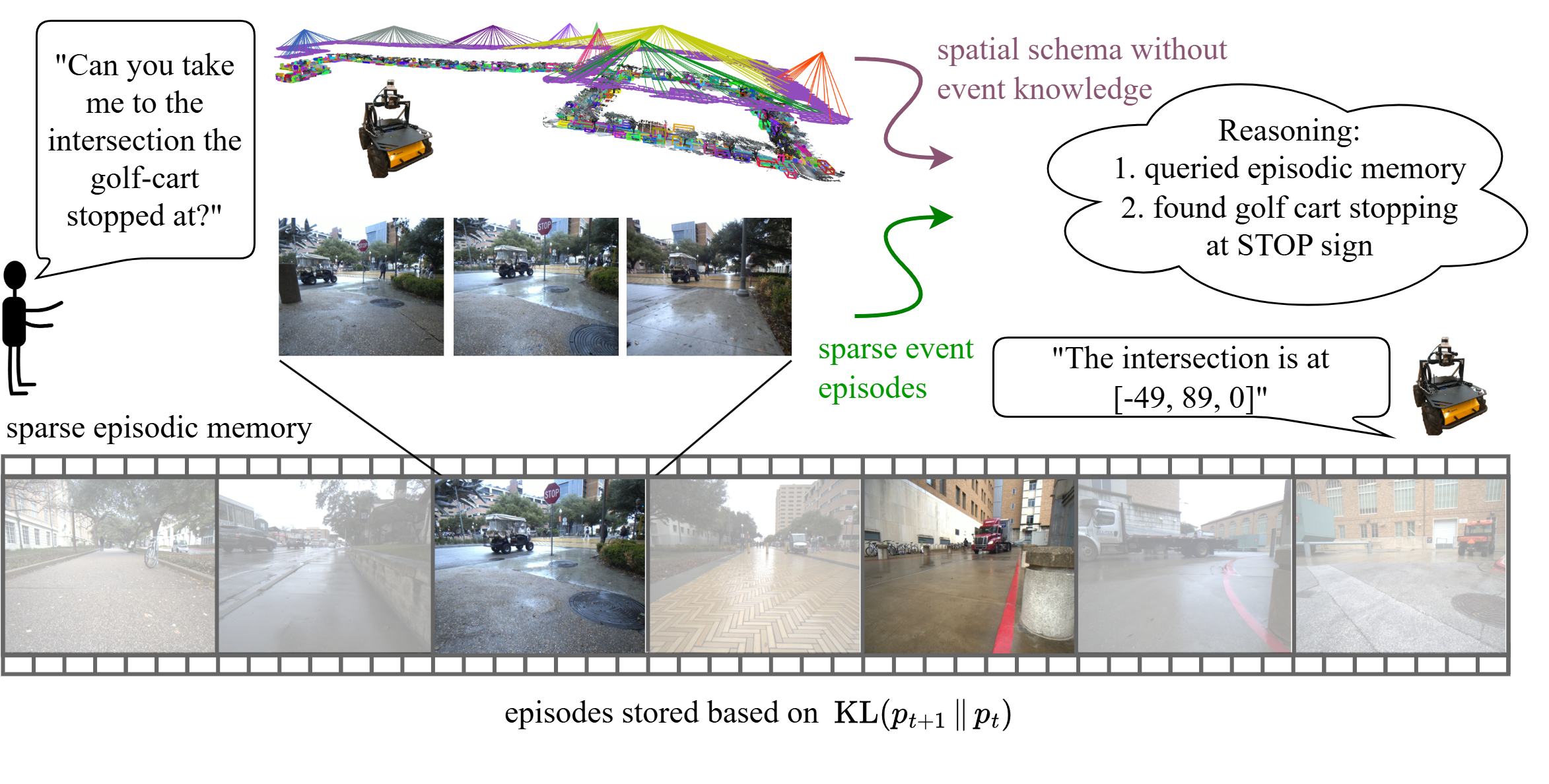}
    \vspace{-1em}
    \captionof{figure}{We present an episodic memory system for robotics that selectively stores high utility experiences as episodic memories via Bayesian surprise of V-JEPA-2 latent space embeddings, to augment 4D scene graph representations. Our system selectively stores sparse episodes during long-horizon ($>$\SI{30}{\minute}) robot rollouts that can later be retrieved by an embodied agent to answer queries and fulfill tasks. 
    }\label{fig:title_figure}
\end{figure}

\section{Introduction}
\label{sec:introduction}
\raggedbottom
Robots operating over extended time horizons may be given tasks or queries requiring knowledge of past events, potentially \textit{far} past, rather than just the current state of the environment. Consider the query: ``Can you take me to the intersection where the golf cart stopped?'' (See~\cref{fig:title_figure}). This requires the robot to remember not only \emph{what exists} but \emph{what happened}. Recalling ``what happened'' requires episodic memory: the ability to retrieve records of specific events that occurred at a particular time and/or place~\cite{Tulving83-elementsEpisodicMemory}.

Current spatio-temporal memory systems for robots, whether based on 3D scene graphs~\cite{Hughes24ijrr-hydraFoundations,Schmid24rss-khronos,Gorlo26cvpr-DAAAM,Gu24icra-conceptgraphs} or view-based retrieval~\cite{Anwar25icra-remembr,Xie24arxiv-EmbodiedRAG}, represent the persistent state of the environment but do not maintain structured records of events.
Recent work has begun to add episodic storage to embodied AI systems~\cite{Ginting25corl-mindpalace,Hu25arxiv-3dllmMem,Shi26iclr-memoryVLA,Lin25arxiv-echoVLA,Baermann25humanoids-hierarchicalEMV,Zhang25arxiv-ella}, but has a common limitation: episodes are stored at fixed temporal intervals, on task-specific triggers, or indiscriminately for all observations.
No principled criterion distinguishes observations worth remembering from redundant ones.
A robot exploring for hours at a fixed storage rate accumulates thousands of episodes, most recording nothing new, while a brief but important event may fall between storage intervals. For robot agents operating in the wild, efficient memory systems need to conserve memory resources by determining which observations are worth remembering with how much detail. 

We can take inspiration from mammalian memory systems that solve this selection problem in part through prediction error.
The Complementary Learning Systems (CLS) framework~\cite{McClelland95pr-CLS,Kumaran16tics-CLSupdated} proposes that the brain maintains two memory systems with fairly distinct roles: a hippocampal system for rapid encoding of specific episodes and a neocortical system for slow extraction of statistical regularities.
Within this architecture, prediction errors inform the encoding hippocampal episodic memory: the hippocampus signals mismatches between model predictions and observations~\cite{Kumaran07jneuro-matchMismatch}. These prediction errors drive updating of related episodic memories~\cite{Sinclair21pnas-predictionError}.
Likewise, predictive coding~\cite{RaoBallard99nn-predictiveCoding,Friston10nrn-freeEnergy} posits that the human brain maintains an expectation $p(z_t \mid z_{<t})$ over upcoming sensory input and reacts to mismatches between expectation and observation.
Event Segmentation Theory~\cite{Zacks07pb-eventSegmentation} similarly proposes that humans segment continuous experience into discrete events at moments with high prediction error. Recent work also argues that prediction error plays a key role in event segmentation~\cite{Nolden24neubiorev-predErrorEventSeg}.
Empirical evidence~\cite{Itti09vr-bayesianSurprise,Kumar23cogsci-bayesSurpriseEventSeg} suggests that a good formalization of this error signal is \emph{Bayesian surprise}, defined as the KL divergence $D_{\mathrm{KL}}(\text{posterior} \| \text{prior})$ over a model space~\cite{Itti09vr-bayesianSurprise} (see \cref{sec:approach-episodic} for more details). It further predicts where humans perceive event boundaries during narrative comprehension~\cite{Kumar23cogsci-bayesSurpriseEventSeg}.
This indicates that the shift in the predictive distribution induced by an observation, rather than the (un-)expectedness of the observation itself, is a good signal to highlight memorable events.
A closely related surprisal-based criterion has further been used to drive online fragmentation of spatial maps in models of grid- and place-cell remapping~\cite{Klukas21biorxiv-fragmentedSpatialMaps}.

Inspired by these findings, we propose Bayesian surprise as a principled gating mechanism for robotic episodic memory formation.
We use a Video Joint Embedding Predictive Architecture (V-JEPA)~\cite{Bardes24tmlr-vjepa,Assran25arxiv-vjepa2} to embed robot observations into a compressed spatio-temporal embedding space, capturing the spatio-temporal continuity of the scene.
We then fit a Gaussian model on past observations as a local predictive model and measure surprise as the change of the predictive model given a new observation. We gate episodes as tail-events of the surprise metric.

We augment DAAAM~\cite{Gorlo26cvpr-DAAAM}, an existing robot memory system that maintains a 4D scene graph as a spatial memory schema, with an additional layer of surprise-gated episodic records. These store memorable events that are used in referring expressions. Also, a temporal sequence of distinct discrete notable events segments the continuous stream of observations at event boundaries, building a structured record of a robots experience. The episodic records can be retrieved by a downstream agent to solve tasks and answer queries. 
\newpage
In summary, we make the following contributions:
\begin{itemize}
    \item A formalization of surprise-gated episodic memory formation for robotic systems, grounding the ``what and when to store'' decision via a prediction-error method, inspired by evidence from cognitive sciences.
    \item An online (causal) unsupervised architecture that computes Bayesian surprise from robot observations within the latent space of a V-JEPA~\cite{Bardes24tmlr-vjepa,Assran25arxiv-vjepa2} model, storing sparse memorable episodes. The architecture integrates with DAAAM~\cite{Gorlo26cvpr-DAAAM}, a robot spatio-temporal memory system.
    \item An evaluation demonstrating the method's utility for downstream robotics applications (long-horizon QA) as well as unsupervised video understanding, demonstrating a $\geq+12\%$ improvement on all QA metrics on the OC-NaVQA benchmark~\cite{Gorlo26cvpr-DAAAM} and outperforming supervised and offline (noncausal) approaches on Generic Event Boundary Detection (GEBD)~\cite{Shou21iccv-GEBD} on the Kinetics-400~\cite{Kay17arxiv-Kinetics} dataset.
\end{itemize}

\section{Related Work}

\subsection{Spatio-Temporal Robot Memory}

Existing spatio-temporal memory systems for robots primarily build persistent semantic representations of the environment.
3D scene graphs~\cite{Armeni19iccv-3DsceneGraphs,Hughes24ijrr-hydraFoundations,Schmid24rss-khronos,Gu24icra-conceptgraphs} organize the environment as hierarchical graphs of entities with semantic attributes grounded in 3D geometry, and have been adopted for task planning~\cite{Rana23corl-sayplan} and grounding~\cite{Maggio24ral-clio}, embodied question answering~\cite{Saxena25corl-GraphEQA,Ginting25corl-mindpalace}, and mobile manipulation.
Recent extensions address temporal dynamics through spatio-temporal metric-semantic SLAM~\cite{Schmid24rss-khronos}, dynamic open-vocabulary scene graphs~\cite{Yan25ral-DovSG}, and dynamic point cloud memory~\cite{Liu25arxiv-dynamem}.
DAAAM~\cite{Gorlo26cvpr-DAAAM} combines 4D scene graph structure with localized VLM captions in real time, capturing the state of the environment at each timestep.

An orthogonal family of approaches forgoes 3D structure and commits VLM annotations of raw RGB frames to various data structures~\cite{Anwar25icra-remembr,Xie24arxiv-EmbodiedRAG,Yang25cvpr-3dmem} for retrieval-augmented reasoning~\cite{Lewis20neurips-RAG}.
Hybrid approaches combine high-detail view-based annotations with lower-detail 3D structure in indoor environments~\cite{Hu25arxiv-3dllmMem}.

All of these systems capture the persistent state of the environment, and some track how that state changes over time.
However, none maintains a structured record of discrete events (an object appearing, a person performing an action, a scene transition, a flash, etc.) that can be queried independently of the current world state.
Answering a query like ``who dropped the vase?'' requires a memory system that records notable events, not just the current state in the scene graph.

\subsection{Episodic Memory in Embodied AI}

Several recent robotic systems combine spatial representations with episodic records. They differ mainly in how episodes are represented and stored.
Some compresses observations into latent vectors~\cite{Rothfuss18ral-deepEpisodicMemory} or store vision-language-model-generated captions and other robot logs in databases~\cite{Anwar25icra-remembr,Xie24arxiv-EmbodiedRAG,Baermann25humanoids-hierarchicalEMV} for later retrieval.
Others retain more spatial structure by using timestamped scene graph instances~\cite{Ginting25corl-mindpalace,Saxena25corl-GraphEQA}, or visual memory snapshots~\cite{Yang25cvpr-3dmem}.
However, for all these methods, observations are stored or sampled at fixed intervals, with no criterion to select informative episodes.
RONAR~\cite{Wang24arxiv-RONAR} takes a step toward selective storage by identifying narration-worthy moments, but using fixed and task-specific hand-designed heuristics.

Another line of research equips large language and vision-language-action models with explicit memory modules.
Working+Episodic memory architectures store spatial maps alongside token-level or embedding-level episode buffers~\cite{Hu25arxiv-3dllmMem,Lin25arxiv-echoVLA,Zhang25arxiv-ella}, or maintain buffers of visual information or semantic embeddings~\cite{Shi26iclr-memoryVLA}; RoboMME~\cite{Dai26arxiv-RoboMME} provides a benchmark for memory-augmented variants of such generalist policies.
These architectures have greatly improved \emph{how} episodes are represented and retrieved, but the core ``when to store'' decision remains hard-coded (\textit{e.g.}, fixed-rate, or driven by uniform temporal sampling).
We are not aware of any existing system with a general criterion for selecting observations worth adding to long-term robot memory. 

\subsection{Surprise and Novelty-Driven Learning}

Use of prediction error in a learned model as an intrinsic motivation signal dates back to early work on curiosity-driven neural controllers~\cite{Schmidhuber91sab-curiosityBoredom}.
In deep reinforcement learning, prediction error is a common intrinsic exploration reward~\cite{Schmidhuber91sab-curiosityBoredom,Houthooft16neurips-VIME,Pathak17icml-curiosityICM,Burda19iclr-RND}. These approaches all use the resulting signal to shape exploration policy.
Curious Replay~\cite{Kauvar23icml-curiousReplay} applies world-model training loss to prioritize which experiences to replay from a buffer for learning from past experiences.

Several recent systems apply surprise-like signals to tasks adjacent to memory formation, without addressing episodic storage directly.
Kullback-Leibler (KL) divergence bounds between prior and posterior in a world model's latent space can be used to detect distributional shifts at deployment time for anomaly detection~\cite{Zollicoffer25icml-noveltyWorldModels}. 
Here, EM-LLM~\cite{Fountas25iclr-emllm} explores thresholding token-leven surprisal to an LLM's context into variable-length events and compress context. D-Mem~\cite{Song26arxiv-DMEM} uses a reward prediction error signal to determine the utility of language inputs and whether they should be integrated into an actively managed agent knowledge graph~\cite{Xu25arxiv-Amem}.

Our work translates these findings to perceptual memory of embodied agents in the wild, using prediction error (Bayesian surprise) for deciding when to store detailed visual information as episodic memory traces. 

\section{Approach}
\label{sec:approach}

\begin{figure}
    \centering
    \includegraphics[width=\linewidth]{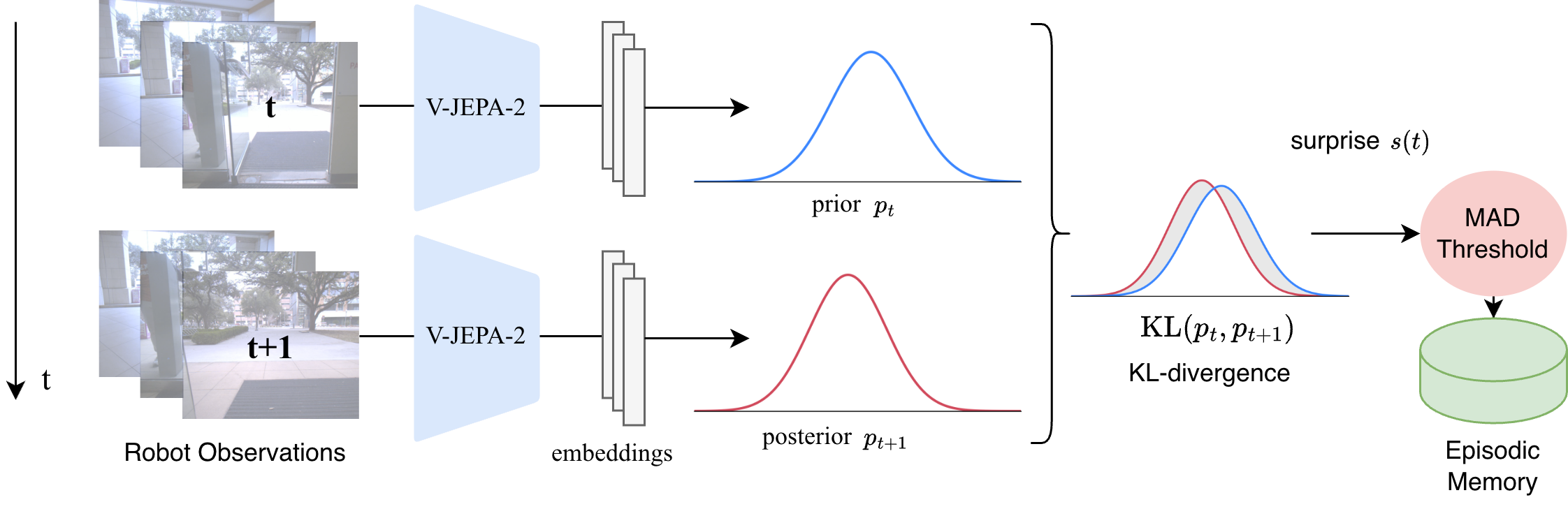}
    \caption{Our proposed pipeline. Two consecutive frame windows are encoded by V-JEPA-2~\cite{Assran25arxiv-vjepa2} into latent features. A diagonal Gaussian fit over past features provides a local predictive distribution, from which we compute the Bayesian KL divergence between consecutive predictives as a per-frame surprise signal. A robust threshold (median absolute deviation) on this signal gates which frames are stored as episodic memory in DAAAM's~\cite{Gorlo26cvpr-DAAAM} scene graph for later agentic retrieval.}
    \label{fig:pipeline}
\end{figure}

We present our approach for gating episodic memories from a robot image sensor stream (see~\cref{fig:pipeline}) based on surprise in a latent V-JEPA-2~\cite{Assran25arxiv-vjepa2} embedding space. 

\subsection{Background: Surprise as Prediction Error}
\label{sec:approach-background}

Inspired by findings in the cognitive sciences literature (see \cref{sec:introduction}) we aim to use Bayesian surprise, the KL divergence between the predictive distribution before and after making a new observation $z_t$ ~\cite{Itti09vr-bayesianSurprise,Kumar23cogsci-bayesSurpriseEventSeg} \emph{as a gating mechanism} for robot episodic memory formation.

To do so, first, surprise must be evaluated in a feature space that ignores stochastic sources of prediction error. For example, if we were to use raw pixel space, tree leaves rustling in the wind would always have high prediction error~\cite{Pathak17icml-curiosityICM}, even though such events are neither surprising nor memorable if one has seen them before. 
Second, the latent space should be as independent as possible of specific deployment, since robot environments and data can vary significantly.
At the current state of research, open-source world models are still often constrained to the video data regimes they were trained on (e.g., Atari, Minecraft, etc.). For in-the-wild robot data, we need the predictive distribution to be available without per-deployment training, as environments can vary significantly.

A Video Joint-Embedding Predictive Architecture (V-JEPA)~\cite{Bardes24tmlr-vjepa,Assran25arxiv-vjepa2} encoder both ignores stochastic sources of error and is trained on large-scale real-world data, therefore being independent of deployment. V-JEPA is self-supervised on a large and general body of unlabeled video data by requiring the model to predict held-out spatio-temporal feature blocks from visible context~\cite{Bardes24tmlr-vjepa,Assran25arxiv-vjepa2}. Through its training, the encoder is shaped to retain only the structure that supports prediction (motion, semantic continuity, scene layout). The resulting features are predictive by construction even though the encoder itself runs as a pure embedding step at inference.

\subsection{Episodic Memory Triggering}
\label{sec:approach-episodic}

To highlight memory episodes, we derive a surprise signal from V-JEPA-2. Transient spikes of this signal would then correspond to episodes that we want to store (see~\cref{fig:pipeline}).

At each timestep, we apply the V-JEPA-2 encoder to a (causal) sliding window of the $M$ most recent frames, pooling its output tokens into a single feature $z_t \in \mathbb{R}^D$ ($D=1024$). Two adjacent windows share $M-1$ of their $M$ ($=64$) frames, so the only difference between $z_{t-1}$ and $z_t$ is a new arriving frame; a shift in $z_t$ therefore tracks the arrival of content that was not predictable from the preceding context.

We propose to use a diagonal Gaussian $p_t = \prod_d \mathcal{N}(\mu_W[t,d], \sigma_W^2[t,d])$, with $d=1,2,\dots,D$, on a sliding window of length $W$ ($=64$) over past values of $z$ as a local predictive distribution, with $\mu_W$ and $\sigma_W$ the per-dimension running mean and standard deviation, respectively. The per-frame Bayesian surprise of $z_t$ under $p_t$, in its $\ell_1$ surrogate form, is
\begin{equation}
    s(t) = \frac{1}{D} \sum_{d=1}^{D}
                  \frac{|z_t[d] - \mu_W[t,d]|}{\sigma_W[t,d]} .
    \label{eq:per-dim-zscore}
\end{equation}

In this Gaussian regime, $s(t)$ tracks the per-frame Bayesian KL divergence between consecutive predictive distributions $p_t$ and $p_{t+1}$. To leading order in $1/W$ (see Appendix \ref{appendix:kl}), the closed-form Gaussian KL divergence reduces to
\begin{equation}
    D_{\mathrm{KL}}(p_{t+1} \,\|\, p_t)
    \approx \frac{1}{2 W^2} \sum_{d=1}^{D}
                \frac{(z_t[d] - \mu_W[t,d])^2}{\sigma_W[t,d]^2}
    = \frac{S_t - C_t}{W^2} ,
    \label{eq:kl-surprise}
\end{equation}
where $S_t = -\log p_t(z_t)$ is the pointwise surprisal of $z_t$ under $p_t$ and $C_t = \tfrac{1}{2}\sum_d \log(2\pi\,\sigma_W[t,d]^2)$ is the Gaussian-entropy term. Note that $C_t$ is independent of $z_t$ and is absorbed by the median absolute deviation (MAD) threshold below. Therefore, in our case of a Gaussian model, surprisal and Bayesian surprise (KL divergence between a prior and posterior distribution) are equivalent up to a $z_t$-independent constant.\footnote{Note that for more powerful predictive families Bayesian surprise has been shown to outperform surprisal in human event segmentation~\cite{Kumar23cogsci-bayesSurpriseEventSeg}.}

Triggering episodes then comes down to finding local maxima in $s(t)$. That is, finding frames, where $s(t)$ exceeds a multiple of its standard deviation. We store an episode at every local maximum of $s(t)$ above a robust threshold,
\begin{equation}
    \tau = \mathrm{med}(s) + \gamma \cdot \frac{1}{\Phi^{-1}(0.75)} \cdot \mathrm{MAD}(s) ,
\end{equation}
where $med(\cdot)$ is the median and $\gamma=1.0$ is a sensitivity threshold. As a final step we perform non-maximum suppression (NMS) to deduplicate clustered peaks. $1/\Phi^{-1}(0.75) \approx 1.4826$ is the Gaussian-consistency factor that makes $1.4826 \cdot \mathrm{MAD}$ an unbiased estimate of the standard deviation under a Gaussian likelihood robust to heavy tails~\cite{Hampel74jasa-influenceCurve,Huber81}. Three parts of our pipeline (V-JEPA's $\ell_1$ training loss~\cite{Bardes24tmlr-vjepa}, the $\ell_1$ surrogate score in \eqref{eq:per-dim-zscore}, and the median$+\gamma \cdot \mathrm{MAD}$ threshold) belong to the same M-estimator family (the $\ell_1$ loss) and are therefore stable in the heavy tails of high-dimensional feature trajectories. Using a squared $\ell_2$-norm instead would amplify tail outliers for each stage.

Each surviving peak emits an event with the trigger timestamp and the robot's pose at that instant. The stored episode is a short window of frames around the trigger together with the surprise score and a joint vision-text embedding for retrieval (\cref{sec:approach-retrieval}).

\subsection{Agentic Retrieval}
\label{sec:approach-retrieval}

We add surprising episodes as an additional layer to DAAAM's~\cite{Gorlo26cvpr-DAAAM} scene graph representation. We store 8-frame episodes for each memory and encode each frame using the constrastive text-image embedding model Perception-Encoder (PE-Core-L14)~\cite{Bolya25neurips-perceptionEncoder}. After mapping, DAAAM's LLM-agent can choose retrieve visual episodes based on the maximum text-image similarity (among the eight frames) to a query, location, or observation time. Retrieved episodes are provided in image-format to the multimodal agent, aiding its reasoning. The episodes augment DAAAM's scene graph as additional data that the LLM can retrieve.

\section{Experiments}

We evaluate the proposed memory formation model in two tasks: Spatio-temporal Robot Question answering (~\cref{sec:exp_qa}), and generic event boundary detection (~\cref{sec:exp_gebd}).

\subsection{Large-scale, long-term spatio-temporal Question Answering (QA)}
\label{sec:exp_qa}

To test the utility of surprise-gated episodic memories for downstream robotics applications, we evaluate our system alongside DAAAM~\cite{Gorlo26cvpr-DAAAM} on spatio-temporal question answering on the challenging OC-NaVQA dataset~\cite{Gorlo26cvpr-DAAAM,Anwar25icra-remembr}. The dataset consists of 7 sequences of in-the-wild recordings of the CODa dataset~\cite{Zhang23arxiv-CODa} and questions referring to any part of the robot's experience and observations along the recorded sequence. Overall, we improve Question accuracy by \textbf{12.0\%}, decrease 
positional error by \textbf{12.4\%}, and temporal error by \textbf{15.7\%}.

\begin{table}[h]
\caption{Ablation on OC-NaVQA dataset~\cite{Gorlo26cvpr-DAAAM}. All models use GPT-5-mini to reason over the constructed memory. ``DAAAM + uniform episodic memory (EM)'' stores observations at fixed intervals and ``+ random episodic memory (EM)'' at random moments instead of using surprise to select which observations to store. Best is \textbf{bold}.
We give each baseline the same episodic memory budget as Ours.}
\label{tab:ocnavqa_ablation}
\centering
\adjustbox{width=.8\columnwidth}{%
\begin{tabular}{lccc}
    \toprule
    Method & \makecell[l]{Question\\ Accuracy $\uparrow$} & \makecell[l]{Positional\\ Error [m] $\downarrow$} & \makecell[l]{Temporal\\ Error [min] $\downarrow$} \\
    \midrule
    DAAAM~\cite{Gorlo26cvpr-DAAAM} & 0.711 & 41.75 & 1.792 \\
    DAAAM~\cite{Gorlo26cvpr-DAAAM} + uniform EM & 0.761 & 40.34 & 1.939 \\
    DAAAM~\cite{Gorlo26cvpr-DAAAM} + random EM & 0.761 & 41.26 & 1.974 \\
    \arrayrulecolor{lightgray} \cmidrule(l{.7em}){1-4} \arrayrulecolor{black}
    \emph{DAAAM + surprise EM (Ours)} & \textbf{0.796} & \textbf{36.57} & \textbf{1.510} \\
    \bottomrule
\end{tabular}}
\end{table}

\begin{table}[h]
\caption{Results on OC-NaVQA dataset~\cite{Gorlo26cvpr-DAAAM}. All models use GPT-5-mini to reason over the constructed memory. $\times$~indicates that method does not keep track of observation times.}
\label{tab:ocnavqa_results}
\centering
\adjustbox{width=.8\columnwidth}{%
\begin{tabular}{lccc}
    \toprule
    Method & \makecell[l]{Question\\ Accuracy $\uparrow$} & \makecell[l]{Positional\\ Error [m] $\downarrow$} & \makecell[l]{Temporal\\ Error [min] $\downarrow$} \\
    \midrule
    ReMEmbR - NVILA-Lite-2B~\cite{Anwar25icra-remembr,Liu2024arxiv-nvila} & 0.432 & 53.466 & 2.287 \\
    ReMEmbR - NVILA-Lite-8B~\cite{Anwar25icra-remembr,Liu2024arxiv-nvila} & 0.463 & 55.894 & 4.106 \\
    \arrayrulecolor{lightgray} \cmidrule(l{.7em}){1-4} \arrayrulecolor{black}
    Concept-Graphs~\cite{Gu24icra-conceptgraphs} & 0.299 & 111.29 & $\times$ \\
    \arrayrulecolor{lightgray} \cmidrule(l{.7em}){1-4} \arrayrulecolor{black}
    DAAAM~\cite{Gorlo26cvpr-DAAAM} & 0.711 & 41.75 & 1.792 \\
    \arrayrulecolor{lightgray} \cmidrule(l{.7em}){1-4} \arrayrulecolor{black}
    \emph{DAAAM + surprise EM (Ours)} & \textbf{0.796} & \textbf{36.57} & \textbf{1.510} \\
    \bottomrule
\end{tabular}}
\end{table}

Even though the dataset does not specifically reference surprising events (\textit{e.g.,} ``Which hand was the person using to pull the cart?''\footnote{examples from benchmark questions}), surprise-gated memories outperform other memory gating mechanisms on average (see~\cref{tab:ocnavqa_ablation}). This is an intuitive observation, as human referring expressions are generally more likely to refer to surprising events.
Uniform or random episodic memory help in distinguishing binary questions that refer to episode-wide information (\textit{e.g.,} ``Is the weather sunny?''\footnotemark[\value{footnote}]), but degrade performance on some temporal and spatial queries where redundant visual information complicates agentic reasoning (See~\cref{tab:ocnavqa_ablation}).

Surprise-gated memory helps in retaining high-resolution visual detail for questions demanding high detail (\textit{e.g.,} ``What was the color of the hat of the person who opened the door for you when you entered the building?''\footnotemark[\value{footnote}]). The system captures the moment of the person opening the door as a surprising event and stores a visual episode, which is retrieved to answer the question.
Note that all methods for  \cref{tab:ocnavqa_ablation} and \cref{tab:ocnavqa_results} are run for three seeds and all results are averaged to remove randomness induced by the use of a Large Language Model as the agent retrieving from the memory.

With $\gamma=1.0$, our method saves on average $1.28$ episodes per minute, resulting in approximately 30 episodes per CODa~\cite{Zhang23arxiv-CODa} sequence and $1.7\%$ of all frames stored.

We also acknowledge that while the episodic memories add useful information for the question answering task, they (a) increase token usage on average by 13\% and (b) require a multimodal LLM for reasoning over the robot memory.

\subsection{Generic Event Boundary Detection}
\label{sec:exp_gebd}

To evaluate our method's generality and capacity to segment videos at generically useful (context shifting) events, we test it on the Generic Event Boundary Detection (GEBD) benchmark~\cite{Shou21iccv-GEBD}.
The goal of the benchmark is to identify natural taxonomy-free boundaries in videos. The canonical evaluation is F1-score measured against a temporal distance threshold relative to the length of the segment.
Our method achieves state-of-the-art results on the Kinetics-400 validation set~\cite{Kay17arxiv-Kinetics}, outperforming prior offline (noncausal) and supervised approaches (see ~\cref{tab:gebd-val-f1}). Typically, GEBD is further evaluated on the TAPOS~\cite{Shao20cvpr-tapos} dataset (an exception to this are other published unsupervised event boundary detection methods~\cite{Wang24tnnls-CoSeg}). As the TAPOS dataset consists of Olympic sports videos, it has a strong domain bias. In addition, event boundaries are semantically informed by knowledge about a particular sport discipline rather than generic. For transparency, we nevertheless present results on the TAPOS~\cite{Shao20cvpr-tapos} dataset in~\cref{appendix:tapos}. 
For very small relative temporal distance thresholds ($0.05-0.10$), offline methods perform slightly better, as being able to look ahead helps them to determine the exact timing of an event boundary.

\begin{table*}[ht]
\centering
\small
\setlength{\tabcolsep}{4pt}
\caption{$F1$-score on Kinetics-GEBD~\cite{Shou21iccv-GEBD,Kay17arxiv-Kinetics} validation set. ``Supervised'': uses GEBD boundary labels at training. ``Online'': causal pipeline, no future-frame access at inference. Best is \textbf{bold}. Baseline results from Jung \textit{et al.}~\cite{Jung25iccv-onlineGEBD} and CoSeg~\cite{Wang24tnnls-CoSeg}.}
\vspace{-5pt}
\label{tab:gebd-val-f1}
\adjustbox{width=\textwidth}{%
\begin{tabular}{p{0.2em}llccccccccccc}
\toprule
&&& \multicolumn{10}{c}{Rel.\,Dis.\ threshold $d$} & \\
\cmidrule(lr){4-13}
& Setting / Method && 0.05  & 0.10  & 0.15  & 0.20  & 0.25  & 0.30  & 0.35  & 0.40  & 0.45  & 0.50  & avg \\
\midrule
&\multicolumn{13}{l}{\emph{offline}} \\
\multirow{11}{0.2em}{\rotatebox[origin=c]{90}{Supervised}} & BMN~\cite{Lin19iccv-bmn} && 0.186 & 0.204 & 0.213 & 0.220 & 0.226 & 0.230 & 0.233 & 0.237 & 0.239 & 0.241 & 0.223 \\
&BMN-StartEnd~\cite{Lin19iccv-bmn}&& 0.491 & 0.589 & 0.627 & 0.648 & 0.660 & 0.668 & 0.674 & 0.678 & 0.681 & 0.683 & 0.640 \\
&TCN-TAPOS~\cite{Lea17cvpr-tcn} && 0.464 & 0.560 & 0.602 & 0.628 & 0.645 & 0.659 & 0.669 & 0.676 & 0.682 & 0.687 & 0.627 \\
&TCN~\cite{Lea17cvpr-tcn} && 0.588 & 0.657 & 0.679 & 0.691 & 0.698 & 0.703 & 0.706 & 0.708 & 0.710 & 0.712 & 0.685 \\
&PC~\cite{Shou21iccv-GEBD} && 0.625 & \textbf{0.758} & 0.804 & 0.829 & 0.844 & 0.853 & 0.859 & 0.864 & 0.867 & 0.870 & 0.817 \\
\cmidrule(lr){2-14}
&\multicolumn{13}{l}{\emph{online}} \\
&Sim-On-BC~\cite{Tang22arxiv-SimOn} && 0.461 & 0.534 & 0.579 & 0.610 & 0.633 & 0.651 & 0.664 & 0.675 & 0.685 & 0.692 & 0.618 \\
&Oad-TR-BC~\cite{Wang21iccv-OadTR} && 0.474 & 0.512 & 0.535 & 0.552 & 0.565 & 0.575 & 0.583 & 0.590 & 0.596 & 0.601 & 0.558 \\
&TeSTra-BC~\cite{Zhao22eccv-TeSTra} && 0.438 & 0.488 & 0.521 & 0.545 & 0.564 & 0.580 & 0.593 & 0.604 & 0.614 & 0.622 & 0.557 \\
&MiniROAD-BC~\cite{An23iccv-MiniROAD} && 0.569 & 0.622 & 0.649 & 0.675 & 0.691 & 0.704 & 0.714 & 0.722 & 0.729 & 0.735 & 0.681 \\
&ESTimator~\cite{Jung25iccv-onlineGEBD} && 0.620 & 0.687 & 0.724 & 0.746 & 0.762 & 0.774 & 0.782 & 0.789 & 0.795 & 0.799 & 0.748 \\
\midrule
&\multicolumn{13}{l}{\emph{offline}} \\
\multirow{6}{0.2em}{\rotatebox[origin=c]{90}{Unsupervised}} & SceneDetect~\cite{Shou21iccv-GEBD} && 0.275 & 0.300 & 0.312 & 0.319 & 0.324 & 0.327 & 0.330 & 0.332 & 0.334 & 0.335 & 0.318 \\
&PA-Random~\cite{Shou21iccv-GEBD} && 0.336 & 0.435 & 0.484 & 0.512 & 0.529 & 0.541 & 0.548 & 0.554 & 0.558 & 0.561 & 0.506 \\
&PA~\cite{Shou21iccv-GEBD}&& 0.396 & 0.488 & 0.520 & 0.534 & 0.544 & 0.550 & 0.555 & 0.558 & 0.561 & 0.564 & 0.527 \\
&CoSeg~\cite{Wang24tnnls-CoSeg} && \textbf{0.656} & \textbf{0.758} & 0.783 & 0.794 & 0.799 & 0.803 & 0.804 & 0.806 & 0.807 & 0.809 & 0.782 \\
\cmidrule(lr){2-14}
&\multicolumn{13}{l}{\emph{online}} \\
&\textbf{Ours} && 0.612 & 0.742 & \textbf{0.806} & \textbf{0.846} & \textbf{0.869} & \textbf{0.881} & \textbf{0.888} & \textbf{0.893} & \textbf{0.897} & \textbf{0.901} & \textbf{0.833} \\
\bottomrule
\end{tabular}
}
\vspace{-5pt}
\end{table*}

\section{Limitations}

While our method is shown to be effective in robotics-relevant settings, we identify several limitations and avenues for improvement.
V-JEPA-2~\cite{Assran25arxiv-vjepa2} is trained in a self-supervised fashion for non-causal prediction. Therefore, we use a Gaussian model as a surrogate causal predictive model. This model is less expressive than a large causal prediction model trained on internet-scale data. We therefore imagine video world model architectures like V-JEPA-2-AC~\cite{Assran25arxiv-vjepa2} to be an even better signal for gating episodic memory storage. 
Additionally, we currently use text-image embeddings for retrieving memory episodes in downstream tasks and store sparse episodes as visual frames. To capture the temporal evolution of events, text-to-video embeddings or different retrieval mechanisms are promising directions of future work. Further, encoding frames in a compressed representation and using replay of episodes to find statistical regularities in stored episodes and distill context-free knowledge is of interest. 

Finally, as we use a sliding-window predictive model, surprising, yet re-appearing subjects will continue to trigger event storage. To this end, habituation mechanisms could be explored in future work.

\section{Conclusion}
We propose a robotic episodic memory framework gating on surprise (broadly recognized in cognitive science as a signal for episodic encoding) computed in the V-JEPA-2~\cite{Assran25arxiv-vjepa2} latent space. The resulting gate is causal, unsupervised and augments the 4D scene graph robot memory DAAAM~\cite{Gorlo26cvpr-DAAAM} with a layer storing generically useful events. Surprise-gated episodes give a $\geq+12\%$ improvement over DAAAM on long-horizon spatio-temporal question answering, and also outperform uniform and random episodic-memory baselines at the same memory budget. The same per-frame scoring method also improves on supervised and offline (noncausal) baselines at Kinetics-400 generic event boundary prediction~\cite{Shou21iccv-GEBD,Kay17arxiv-Kinetics} while running unsupervised and online.

\clearpage

\bibliography{../references/myRefs,../references/refs,references/new_refs}  %
\newpage
\appendix
\crefalias{section}{appsec}
\section{Bayesian KL Divergence as Surprisal for the Sliding Diagonal Gaussian}
\label{appendix:kl}

We derive \cref{eq:kl-surprise}, showing the per-frame Bayesian KL divergence between consecutive sliding-window Gaussian distributions approximately reduces to the pointwise surprisal of $z_t$ under $p_t$.

For two univariate Gaussians $\mathcal{N}(\mu_0, \sigma_0^2)$ and
$\mathcal{N}(\mu_1, \sigma_1^2)$,
\begin{equation}
    D_{\mathrm{KL}}\bigl(\mathcal{N}(\mu_1, \sigma_1^2) \,\|\, \mathcal{N}(\mu_0, \sigma_0^2)\bigr)
    = \log\frac{\sigma_0}{\sigma_1}
        + \frac{\sigma_1^2 + (\mu_1 - \mu_0)^2}{2\sigma_0^2}
        - \frac{1}{2} .
    \label{eq:closed-form-kl}
\end{equation}
Let $\Delta\mu = \mu_1 - \mu_0$ and $\Delta\sigma = \sigma_1 - \sigma_0$.
Taylor-expanding $\log(\sigma_0/\sigma_1) = -\log(1 + \Delta\sigma/\sigma_0)$ about $\Delta\sigma = 0$ gives $-\Delta\sigma/\sigma_0 + \tfrac{1}{2}(\Delta\sigma/\sigma_0)^2 + O((\Delta\sigma/\sigma_0)^3)$, while $\sigma_1^2/(2\sigma_0^2) - \tfrac{1}{2}$ expands exactly to $\Delta\sigma/\sigma_0 + \tfrac{1}{2}(\Delta\sigma/\sigma_0)^2$. The linear-in-$\Delta\sigma$ contributions of the two terms cancel, leaving the KL divergence of the $d$'th dimension as
\begin{equation}
    D_{\mathrm{KL}, d} = \frac{(\Delta\mu)^2}{2 \sigma_0^2}
                       + O\bigl((\Delta\sigma/\sigma_0)^{2}\bigr) .
    \label{eq:perdim-kl-expanded}
\end{equation}

Incorporating the new observation $z_t$ drops $z_{t-W}$ from the window, so the exact mean update is
\begin{equation}
    \Delta\mu_W[t,d]
    = \frac{z_t[d] - \mu_W[t,d]}{W}
    \;+\; \frac{\mu_W[t,d] - z_{t-W,d}}{W} .
\end{equation}
The second term here depends on the dropped $z_{t-W}$ and not on $z_t$, and is collected into the $O(D/W^2)$ residual below. The first-order update of the running standard deviation is
\begin{equation}
    \Delta\sigma_W[t,d]
    = \frac{1}{2 W} \cdot
        \frac{(z_t[d] - \mu_W[t,d])^2 - \sigma_W[t,d]^2}{\sigma_W[t,d]}
        + O(1/W^2) .
\end{equation}

\paragraph{Aggregated identity.}
Substituting both updates into~\cref{eq:perdim-kl-expanded} and
summing over the $D$ dimensions,
\begin{equation}
    D_{\mathrm{KL}}(p_{t+1} \,\|\, p_t)
    = \frac{1}{2 W^2} \sum_{d=1}^{D}
            \frac{(z_t[d] - \mu_W[t,d])^2}{\sigma_W[t,d]^2}
        + O(D/W^2) ,
\end{equation}
the $O(D/W^2)$ residual contains variance-shift term and mean-substitution error. This is~\cref{eq:kl-surprise}.

\newpage 
\section{TAPOS GEBD evaluation}
\label{appendix:tapos}

GEBD~\cite{Shou21iccv-GEBD} has historically been evaluated on both the Kinetics-400 dataset~\cite{Kay17arxiv-Kinetics} and the TAPOS~\cite{Shao20cvpr-tapos} dataset. An exception to this is CoSeg~\cite{Wang24tnnls-CoSeg}, likewise an unsupervised method.

In~\cref{tab:tapos-val-f1} we present the results of our method on the TAPOS dataset~\cite{Shao20cvpr-tapos}, where notably our method is outperformed by other approaches. This is due to several factors: First, TAPOS focuses on one particular domain (videos of Olympic events). Thus, supervision helps a lot to adapt to the domain. Additionally, event boundaries are often semantic rather than generic (e.g., boundaries in between ``running'' to ``jumping'' to ``landing'' in a video of a triple-jump event), rendering online surprise less effective of an event boundary predictor and also rewarding methods that can look ahead and determine baseline surprise based on the future. 

\begin{table*}[ht]
\centering
\small
\setlength{\tabcolsep}{4pt}
\caption{$F1$-score on TAPOS~\cite{Shao20cvpr-tapos} validation set. Baseline results from Jung \textit{et al.}~\cite{Jung25iccv-onlineGEBD}.}
\label{tab:tapos-val-f1}
\adjustbox{width=\textwidth}{%
\begin{tabular}{p{0.2em}llccccccccccc}
\toprule
&&& \multicolumn{10}{c}{Rel.\,Dis.\ threshold $d$} & \\
\cmidrule(lr){4-13}
& Setting / Method && 0.05  & 0.10  & 0.15  & 0.20  & 0.25  & 0.30  & 0.35  & 0.40  & 0.45  & 0.50  & avg \\
\midrule
&\multicolumn{13}{l}{\emph{offline}} \\
\multirow{11}{0.2em}{\rotatebox[origin=c]{90}{Supervised}} & ISBA~\cite{Shou21iccv-GEBD} && 0.106 & 0.170 & 0.227 & 0.265 & 0.298 & 0.326 & 0.348 & 0.348 & 0.348 & 0.348 & 0.330 \\
& TCN~\cite{Lea17cvpr-tcn} && 0.237 & 0.312 & 0.331 & 0.339 & 0.342 & 0.344 & 0.347 & 0.348 & 0.348 & 0.348 & 0.330 \\
& CTM~\cite{Shou21iccv-GEBD} && 0.244 & 0.312 & 0.336 & 0.351 & 0.361 & 0.369 & 0.374 & 0.381 & 0.383 & 0.385 & 0.350 \\
& TransParser~\cite{Shao20cvpr-tapos} && 0.289 & 0.381 & 0.435 & 0.475 & 0.500 & 0.514 & 0.527 & 0.534 & 0.540 & 0.545 & 0.474 \\
& PC~\cite{Shou21iccv-GEBD} && \textbf{0.522} & \textbf{0.595} & \textbf{0.628} & \textbf{0.646} & \textbf{0.659} & \textbf{0.665} & \textbf{0.671} & \textbf{0.676} & \textbf{0.679} & \textbf{0.683} & \textbf{0.642} \\
\cmidrule(lr){2-14}
&\multicolumn{13}{l}{\emph{online}} \\
& Sim-On-BC~\cite{Jung25iccv-onlineGEBD} && 0.225 & 0.269 & 0.303 & 0.329 & 0.350 & 0.367 & 0.381 & 0.394 & 0.405 & 0.415 & 0.344 \\
& Oad-TR-BC~\cite{Jung25iccv-onlineGEBD} && 0.263 & 0.319 & 0.361 & 0.394 & 0.422 & 0.445 & 0.465 & 0.483 & 0.497 & 0.510 & 0.416 \\
& TeSTra-BC~\cite{Jung25iccv-onlineGEBD} && 0.364 & 0.417 & 0.452 & 0.478 & 0.496 & 0.511 & 0.523 & 0.533 & 0.542 & 0.550 & 0.487 \\
& MiniROAD-BC~\cite{Jung25iccv-onlineGEBD} && 0.422 & 0.472 & 0.502 & 0.522 & 0.537 & 0.549 & 0.558 & 0.566 & 0.572 & 0.578 & 0.528 \\
& ESTimator~\cite{Jung25iccv-onlineGEBD} && 0.394 & 0.455 & 0.499 & 0.532 & 0.558 & 0.578 & 0.594 & 0.608 & 0.619 & 0.629 & 0.547 \\
\midrule
&\multicolumn{13}{l}{\emph{offline}} \\
\multirow{5}{0.2em}{\rotatebox[origin=c]{90}{Unsupervised}} & SceneDetect~\cite{Shou21iccv-GEBD} && 0.035 & 0.045 & 0.047 & 0.051 & 0.053 & 0.054 & 0.055 & 0.056 & 0.057 & 0.058 & 0.051 \\
& PA-Random~\cite{Shou21iccv-GEBD} && 0.158 & 0.233 & 0.273 & 0.310 & 0.331 & 0.347 & 0.357 & 0.369 & 0.376 & 0.384 & 0.314 \\
& PA~\cite{Shou21iccv-GEBD} && 0.360 & 0.459 & 0.507 & 0.543 & 0.567 & 0.579 & 0.592 & 0.601 & 0.609 & 0.615 & 0.543 \\
\cmidrule(lr){2-14}
&\multicolumn{13}{l}{\emph{online}} \\
& \textbf{Ours}\textsuperscript{*} && 0.260 & 0.336 & 0.379 & 0.413 & 0.437 & 0.458 & 0.474 & 0.485 & 0.493 & 0.499 & 0.423 \\
\bottomrule
\end{tabular}
}
\footnotesize{\textsuperscript{*}as per instructions of GEBD~\cite{Shou21iccv-GEBD}, we do not include videos no longer available on YouTube.}
\vspace{-5pt}
\end{table*}

\end{document}